%% file: IEEE_Paper_Example.tex
\documentclass[conference,compsoc]{IEEEtran}

\renewcommand\IEEEkeywordsname{Keywords}

\ifCLASSINFOpdf
  \usepackage[pdftex]{graphicx}
  \graphicspath{ {pictures/}{../pdf/}{../jpeg/} }
  \DeclareGraphicsExtensions{.pdf,.jpeg,.png,.PNG}
\else
\fi
\ifCLASSOPTIONcompsoc
 \usepackage[caption=false,font=normalsize,labelfont=sf,textfont=sf]{subfig}
\else
 \usepackage[caption=false,font=footnotesize]{subfig}
\fi
\usepackage{flushend}
\usepackage{tabularx,booktabs,textcomp}
\newcolumntype{C}{>{\centering\arraybackslash}X} 
\newcolumntype{L}{>{\raggedright\arraybackslash}X} 
\usepackage[noadjust]{cite}

\usepackage{wasysym}
\usepackage{lipsum} 

\usepackage{color}
\PassOptionsToPackage{hyphens}{url}
\usepackage{hyperref}
\begin{document}
\title{Not-in-Perspective: 
Towards Shielding Google's Perspective API Against Adversarial Negation Attacks}


\author{
  \IEEEauthorblockN{Michail Alexiou}
  \IEEEauthorblockA{\textit{Department of Computer Science} \\
    \textit{Kennesaw State University}\\
    Marietta, GA, USA \\
    malexiou@kennesaw.edu}
  \and
  \IEEEauthorblockN{Sukarno J. Mertoguno}
  \IEEEauthorblockA{\textit{School of Cybersecurity and Privacy} \\
  \textit{Georgia Institute of Technology}\\
    Atlanta, GA, USA \\
    karno@gatech.edu}
}


\IEEEoverridecommandlockouts

\maketitle

\IEEEpubidadjcol

\begin{abstract}
The rise of cyberbullying in social media platforms involving toxic comments has escalated the need for effective ways to monitor and moderate online interactions. Existing solutions of automated toxicity detection systems, are based on a machine or deep learning algorithms.
However, statistics-based solutions are generally prone to adversarial attacks that contain logic based modifications such as negation in phrases and sentences. In that regard, we present a set of formal reasoning-based methodologies that wrap around existing machine learning toxicity detection systems. Acting as both pre-processing and post-processing steps, our formal reasoning wrapper helps alleviating the negation attack problems and significantly improves the accuracy and efficacy of toxicity scoring. We evaluate different variations of our wrapper on multiple machine learning models against a negation adversarial dataset. 
Experimental results highlight the improvement of hybrid (formal reasoning and machine-learning) methods against various purely statistical solutions.
\end{abstract}





%



\input{sections/01-intro}

\input{sections/09-RelaWorks}
\input{sections/04-Implementation}
\input{sections/05-Evaluation}
\input{sections/06-Discussion}

\input{sections/08-Conclusions}





\nocite{}
\bibliography{references}
\bibliographystyle{IEEEtran}
\end{document}

%% file: sections/01-intro.tex
\section{Introduction}
\label{s:intro}

In an era of constant exposure to social media platforms, the necessity of developing tools to monitor them and ensure their safe use is more prominent than ever. About 97\% of teenagers ~\cite{mayo-0} and 86\% young adults ~\cite{etactics-0} 
worldwide are considered active users of social media, with the total amount of time spent in online platforms increasing by a 20\% margin ~\cite{security-0} 
during the pandemic. Approximately 38\% of them ~\cite{dataprot-0} have experience online toxicity in social media.
As of 2018, more than 59\% of US teenagers have being exposed to some form of online toxicity ~\cite{dataprot-0}. Cyberbullying may manifest in various forms such as religious hatred, body shaming, attack on sexual orientation and etc., creating an unpleasant environment for social media users and even leading to severe mental health issues ~\cite{etactics-0}. 
So far most of the automatic monitoring efforts have been proven inadequate against 
cyberbullying, with social media slowly being converted into a marketplace where toxicity draws in all the attention of the customers \cite{entrepreneur-0}. Google developed and published Perspective API \cite{WIRED-Greenberg} as a measure to track and counter the growing toxicity in social media platforms, it was quickly proven to be prone to several different classes of adversarial attacks.


Hosseini et. al. \cite{Hosseini-RP3} demonstrated example attacks on Google’s Perspective API using adversarial data as input. Toxic sentences were modified to include the adverb 'not' in front of each offensive word. Even though, the new sentences did not represent the same level of offensive language compared to their original counterparts, Perspective still predicted high toxicity scores.
A published article considering Hosseini's exploits, stated that Google acknowledges its Machine Learning (ML) isn't enough. In its Google's gitHub repository, it is advised to not use the software for automated moderation because "the models make too many errors." ~\cite{TheRegister-Claburn}. 
In the beginning of our research, we confirmed that Google's Perspective API has not significantly improved since the publication of Hosseini et. al. \cite{Hosseini-RP3}. Additional objections about Perspective's capabilities have been raised regarding unintentional classification of non-offensive terms as toxic. Google responded to Perspective’s limitations by making it’s API and datasets available to the public, thus, relying to human annotators for a solution. However, members of Google’s Jigsaw team \cite{jigsaw-0} have argued about the effects of an individual's inherent subjectivity when it comes to identifying a given sentence as toxic or non toxic.

Google's Perspective API is susceptible to adversarial text attacks, which are categorized into (i) typo attacks that leverage the discrepancy between human and machine interpretation of offensive words and phrases with perturbed letters and (ii) negation attacks that exploit the fundamental difficulty of statistics and machine-learning when dealing with the semantic transformation caused by negation.
Both 
classes exploit similar evasion techniques against machine learning. In both case, adversarial training is not practical due to the extremely large number of attack variations for each specific word/phrase. 
Jain et. al. \cite{adv_text} described a framework that can automatically produce adversarial sentences against Perspective, from an unlabeled dataset of Facebook. However, using unlabeled data and relying on Perspective for their initial toxicity scores can lead to underestimating potential adversarial data entries in the original dataset. Additionally, this particular methodology favors small perturbations at sentence level instead of ensuring negation-based adversarial modifications on the sentences. 

In the current paper, we address this issue by introducing a formal reasoning wrapper around learning-based methods. The goal is to enhance the existing ML methods by incorporating heuristic logic, while also preserving the advantages of using a statistical tool as the backbone. 
Over the years, multiple defense strategies have been presented for protection against adversarial \cite{ICLR_adversar} and adaptive \cite{NEURIPS2020_adaptive_attacks} attacks, either based on formal reasoning or machine learning (e.g., adversarial training). However, for the provision of an effective defense against such attacks, one must consider each attack as a unique and individual case. Thus, in the current manuscript we specifically focus on a single adaptive/adversarial attack, which is the purposeful negation of toxic comments to expose the inconsistency of machine learning approach. 
We don’t focus on sarcasm, since it constitutes a subjective process which relies on the reader’s experiences and beliefs. 
The results of extensive experiments support the stated hypothesis, that learning-based solutions are 
vulnerable to adversarial attacks, regardless of the training methods. Additionally, they prove that logic-based adversarial attacks can be addressed by synergistic solutions consisting of heuristic and learning methods. 

%% file: sections/09-RelaWorks.tex
\section{Related Works}
\label{s:relawork}

Rodrigues et. al. \cite{Rodriguez2018ShieldingGL} evaluate Perspective against adversarial attacks (i) based on obfuscation and (ii) based on polarity. Obfuscation modifications are essentially the replacement of certain keywords with counterparts that contain numbers or symbols instead of specific characters. Polarity augmentations include the addition of negation before certain keywords. The authors prove the effectiveness of both types of attacks by creating a combinatorial model TextPatrol API and Google Perspective API which improve the deduced toxicity scores however still produces insufficient result improvements. They concluded that one of the most effective ways to defend against such attacks is to remove the negated part of the sentence altogether. Grondahl et. al. \cite{alllove} test a variety of different models trained for hate speech detection against a wide range of adversarial attacks. Their results indicate that most learning based techniques produce comparable results provided that they follow standard ways of training, even in the case of adversarial training sets. Furthermore, they highlight the effectiveness of additional types of attacks on top of comment negation. They deduce that embedding the word “love” in the input text is the most effective method to decrease the overall toxicity, since it has a low toxicity weight in every dataset.

Jain et. al. \cite{adv_text} propose a methodology for the automatic generation of adversarial dataset that can attack Google’ Perspective API. In particular, a number of preprocessed and unlabeled text data are queried into Perspective and their toxicity scores are retrieved. A second model based on a word-level Convolutional Neural Network is trained using the aforementioned toxicity scores. Finally, the Carlini-Wagner attack \cite{carl-wag} is deployed to deduce the adversarial features capable of confusing Perspective into misclassifying input text. Brassard et. al. \cite{BrassardGourdeau2018ImpactOS} propose counteracting adversarial attacks by integrating sentiment analysis with learning-based toxicity detection models. Their hypothesis dictates that the sentiment embedded in a sentence can't change even though words are altered or masked. These claims are tested by evaluating the method on multiple datasets, however, the results of the toxicity detection model with and without sentiment analysis are similar.

%% file: sections/04-Implementation.tex
\section{Implementation}
\label{s:implementation}

\subsection{Overall Methodology}
To deal with the problem of negation in adversarial toxic comments, a formal reasoning module is utilized for the detection of phrases affected by negation within a sentence. More specifically, we think of the proposed approach as a pre-processing and post-processing layer, that wraps around a machine learning based toxicity classifier. The sentence’s toxicity score is recomputed with respect to the logical and semantic implications of negation. The input comment is initially parsed through a pre-processing module to replace negation contractions (i.e., “weren’t → were not” and etc). 
Two methods are considered capable of dealing with the concept of negation, which are (1) readjusting the computed toxicity score based on a heuristic formula and (2) substituting the word or words that are negated with their antonyms, extrapolate the set of queried comments based on the number of semantically fitting antonyms and computing the average toxicity score of all the produced sentences. 

As described in previous section, an insertion of negation (not) does not significantly change the toxicity score produced by Perspective API \cite{Hosseini-RP3}, which is incorrect. Negation inverses the toxicity score of the attached word or phrase and significantly affects the overall toxicity. A negation attack is fundamentally difficult for machine learning to address.
Machine learning is essentially a statistical procedure which relies on frequency of occurrence and correlation of terms in its input vector(s) to calculate prediction/decision. Anticipating occurrences of negation or 'not' using adversarial training \cite{adv_text} significantly blows up the size of the training set and may potentially degrade the performance of the neural network (machine learning). The proposed methods for integrating approaches of logical reasoning and statistical machine learning is consistent with the 'Learn2Reason' concept \cite{L2R1}, which argues that certain classes of problems are better solved with formal-reasoning while others require statistical approach. Thus, a synergy between both of them is often required for non-trivial problems. 

\subsection{Reasoning and Synthesis}
\label{ss:ReaSynt}

Heuristic adjustment of toxicity score is based on the concept of treating negation in a sentence 
as logical NOT, which reverses the value of a variable.
Thus, the toxicity score (TS) of a sentence containing logical negation can be represented as “1- TS”. However, not every word in the sentence is affected by negation. 
Stanford’s Part-of-Speech tagger \cite{posTagger-2k11} is used to detect the affected words based on their their location and syntactic meaning via recursive parsing. 
The recursion is extended when we reach tokens of conjunction (i.e. and, or, etc.) and commas but terminates in all other cases (i.e., full-stops, exclamation marks). The distributive mathematical property is also used to deduce the extend of negations effects.

\begin{figure}[ht]
   \centering
    \includegraphics[width=0.45\textwidth]{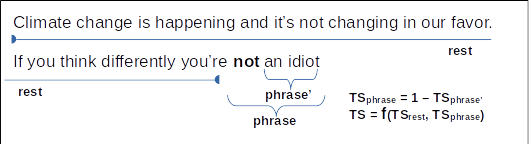}
    \caption{Dealing with negation.}
    \label{fig:NegInPhra}
\end{figure}

The input sentence is divided into segments affected by negation (SN) and segments that aren’t (S). Each segment is queried into a trained classifier and its toxicity is retrieved to compute the total score, which is based on the weighted average of the scores of its S and S\textsubscript{N} segments. The weighted average recalculation of the toxicity, taking logical negation into account, is illustrated in Eq. \ref{eqn:heur11} with the weights being calculated using Eq. \ref{eqn:heurweight}. The weights for each segment are determined based on the segment’s number of words divided by the sentence’s overall number of words. 
In particular, that specific equation represents an equal weight distribution of the overall toxicity among the sentence's words, considering logic negation as an indicator for the recomputation of their individual toxicity. Figure \ref{fig:NegInPhra} shows an example for the aforementioned approach. We diversify our approaches by considering 2 more simple variations of aforementioned equation presented in Eq. \ref{eqn:heur2} and Eq. \ref{eqn:heur3}. More specifically, in Eq. \ref{eqn:heur2} we only use logical negation as the basis for the equation, while in Eq. \ref{eqn:heur3} we merely redistribute the toxicity scores based on the values of the weights.

\begin{equation}
\label{eqn:heur11}
TS=w_1*TS_1 + w_2 * (1 - TS_2) + ... + w_N * (1 - TS_N)
\end{equation}

\begin{equation}
\label{eqn:heurweight}
w_i = S_{i Number of Words} / Sentence_{Number of Words}
\end{equation}

\begin{equation}
\label{eqn:heur2}
TS= TS_1 + (1 - TS_2) + ... + (1 - TS_N)
\end{equation}

\begin{equation}
\label{eqn:heur3}
TS=w_1*TS_1 + w_2 * TS_2 + ... + w_N * TS_N
\end{equation}

We continue to use sentence \textit{“Climate change is ... you are an idiot”} as an example to showcase the workflow of the aforementioned methodologies. More specifically, the input sentence is split into the following segments: 1- “Climate change is happening and it is”, 2- “not changing”, 3- “in our favor. If you think differently you are”, 4- “not an idiot”. From the previous segments, only 2 and 4 contain negation. However, segment 2 doesn’t contain any offensive term and, therefore, processing it based on logical negation increases the toxicity. 
To overcome this issue, we retrieve the toxicity of the phrase from the corresponding toxicity detection model and if it’s over 0.5 then “not” is treated as logical negation. The individual toxicity scores for segments S\textsubscript{2} and S\textsubscript{4} are TS\textsubscript{2} = 0.034 and TS\textsubscript{4} = 0.932 respectively, thus, only the latter segment is processed. Similarly, we compute TS\textsubscript{1} = 0.068 and TS\textsubscript{3} = 0.12. The corresponding weights for each segment are computed accordingly based on the number of words in each segment with regard to the total number of words of the input sentence. After applying the equations \ref{eqn:heur11}, \ref{eqn:heur2} and \ref{eqn:heur3} we approximately get the toxicity scores 0.107, 0.198 and 0.333 respectively.

Additional methods for combining segments of toxicity scores into overall score are achieved by approximating the weights machine-learning (e.g. Perspective) assigned to each segments. 
Let's consider a compound of sentences shown in figure \ref{fig:NegInPhra}: \textbf{A} - \textit{“Climate change is happening and it’s not changing in our favor. If you think differently you are \textbf{not} an idiot”} and \textbf{A'} - \textit{“Climate change is happening and it’s not changing in our favor. If you think differently you are an idiot”}.
Let's consider \textbf{B} be the common segments of \textbf{A} and \textbf{A'} with \textbf{B} - \textit{“Climate change is happening and it’s not changing in our favor. If you think differently you are”}, while \textbf{C} and \textbf{C'} are \textit{"\textbf{not} an idiot”} and \textit{"an idiot”} respectively. In this case, sentence \textbf{A} is composed of \textbf {B} and \textbf{C}, while \textbf{A'} is composed of \textbf {B} and \textbf{C'}. In this method, \(TS_{A'}, TS_B\) and \(TS_{C'}\) are evaluated from at least three different queries to a machine learning algorithm and we expect that they are evaluated with different factor or 'gain'. Hence, \(TS_{A'} \neq TS_B + TS_{C'}\). Assuming that the machine-learning algorithm is relatively consistent, we can expect that the following equations, 
\[ TS_{A'} = \alpha_1.TS_B + \beta_1.TS_{C'} \quad \textrm{and} \quad TS_A = \alpha_0.TS_B + \beta_0 .TS_C\]
hold true. Similarly, as the two compound of sentences are almost identical (besides negation), we expect that 
\[\alpha_0 \approx \alpha_1 \equiv \alpha \quad \textrm{and} \quad \beta_0 \approx \beta_1 \equiv \beta\]
We know that machine-learning algorithm will have difficulty of predicting \(TS_A\) and \(TS_C\). However, it is expected to correctly evaluates  \(TS_{A'}, TS_B\) and \(TS_{C'}\). With only one equation available,
\begin{equation}
\label{eqn:albe}
    TS_{A'} = \alpha.TS_B + \beta.TS_{C'}
\end{equation} 
we cannot compute the two constants \(\alpha\) and \(\beta\).  
Hence, we approximate equation \ref{eqn:albe} with either
\begin{equation}
\label{eqn:approx0}
    TS_{A'} = TS_B + \zeta.TS_{C'} \quad \textrm{and} \quad
    TS_A = TS_B + \zeta.TS_C
\end{equation} 
assuming the prediction of \(TS_{A'}\) and \(TS_B\) has similar 'gain', or
\begin{equation}
\label{eqn:approx1}
    TS_{A'} = \zeta.(TS_B + TS_{C'}) \quad \textrm{and} \quad
    TS_A = \zeta.(TS_B + TS_C)
\end{equation} 
assuming that the prediction of \(TS_{A'}\) is a 'scaled sum' of prediction for \(TS_B\) and \(TS_{C'}\).
The approximation provided by equation \ref{eqn:approx0} or equation \ref{eqn:approx1}, allows us to compute \(\zeta\) from \(TS_{A'}, TS_B\) and \(TS_{C'}\). Since, \(TS_C\) is \((1 - TS_{C'})\), \(TS_A\) can also be computed from either equations \ref{eqn:approx0} or equations \ref{eqn:approx1}. 

\subsection{Substitution and Exploration}
\label{ss:SubsExpl}

\begin{figure}[ht]
   \centering
    \includegraphics[width=0.45\textwidth]{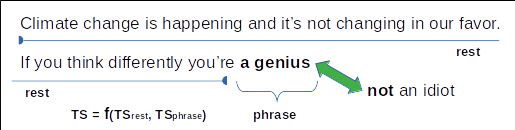}
    \caption{Dealing with negation via replacement.}
    \label{fig:PhraRpl}
\end{figure}

In the current methodology, the words affected by 'not' are replaced by their antonyms, to produce diverse sentences without deviating from the original semantic meaning.
Stanford’s PoS tagger \cite{posTagger-2k11} is utilized for the identification of the words in the sentence affected by negation. Once the negated words are located, a crawler is used to retrieve their antonyms from Thesaurus. 
The retrieved antonyms substitute their corresponding negated words in the original text generating a new sentences. At most 5 antonyms are retrieved for each negated word to acquire sufficient representation without increase in processing time. Then, Parrot \cite{prithivida2021parrot} extrapolates the available number of produced sentences. It is selected for its ability to generate diverse sentences with respect to the original text, while maintaining semantic meaning and structural integrity. It also enables the user to control hyper parameters (e.g., fluency threshold,  the number and length of generated paraphrases, aspiring diversity level) with ease via scripting. Therefore, the overall toxicity of the original sentence is calculated as the average score of all paraphrases produced by Parrot. Figure \ref{fig:PhraRpl} demonstrates an example of the aforementioned approach.

A variation of the aforementioned methodology is also developed, where a crawler retrieve antonyms for the words of interest based on their contextual meaning. The Lesk algorithm \cite{lesk_algo} is used to extract the definition of the word that is embedded in the sentence with respect to the rest of sentence’s text. This step helps to avoid any word-meaning ambiguities and to produce sentences that are closer to the original meaning. Then, the crawler retrieves antonyms from OneLook Reverse Dictionary 
using the context description extracted by the Lesk algorithm. Lastly, the Parrot paraphraser is again used to extrapolate and diversify the number of produced sentences and their average toxicity score is computed. 

\begin{figure}[ht]
   \centering
    \includegraphics[width=0.45\textwidth]{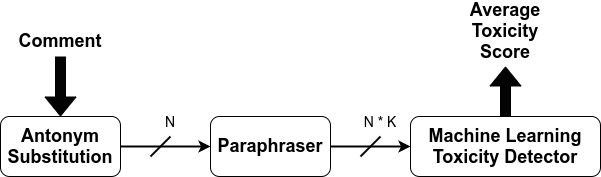}
    \caption{Substitution and Extrapolation's Workflow.}
    \label{fig:Method2 overview}
\end{figure}

The sentence \textit{“Climate change is happening ... you are not an idiot”} is used as the example to illustrate the workflow of the subtraction and extrapolation methods. The workflow diagram of both methodologies including all of it’s processing steps is presented in Fig \ref{fig:Method2 overview}. In particular, N antonyms are retrieved to replace phrases affected by negation and K paraphrases are generated for each combination of antonyms. Contrarily to heuristic adjustment, in this methodology detecting and substituting the phrase “not changing” doesn’t have a major impact in the overall sentence's toxicity, because it doesn’t affect its semantic meaning. Then, Parrot \cite{prithivida2021parrot} receives the substituted sentences and generates a set of paraphrased texts based on them. Finally, the produced paraphrases are queried by an ML toxicity detector, while their scores are retrieved and averaged to compute the overall toxicity of the original sentence. The total number of paraphrases produced by the first version of substitution and extrapolation is 12, with an average toxicity score 0.334. Similarly, the second version results in 25 paraphrases with an average score of 0.286.


%% file: sections/05-Evaluation.tex
\section{Evaluation}
\label{s:eval}

\subsection{Dataset and Models}
We perform an extensive evaluation of the proposed methodologies and a set of baseline models against a negated adversarial dataset. These models are (1) Perspective API \cite{WIRED-Greenberg}, (2) Long short-term memory (LSTM) \cite{hochreiter1997long} model combined with a Convolutional Neural Network (CNN) and (3) Bidirectional Encoder Representations from Transformer (BERT) \cite{Devlin2019BERTPO} on top of an LSTM model. The negated test set is generated from existing toxicity datasets available online. The baseline models are also used in the place of the toxicity detectors within the scope of the methodologies presented in section 4. This step aims to diversify our research approaches and get a better understanding of the overall performance of each solution in direct comparison to each other. All models are trained in a Google Colab Pro+ environment with a GPU hardware accelerator.

\begin{itemize}

\item \textbf{Perspective API:} Perspective API is a tool that assigns toxicity scores to queries based on their contents. It was created by the Google Counter-Abuse Technology team and the Jigsaw research team in order to monitor behavior in social media posts and comments. 
More specifically, it assigns a toxicity score from 0 and 1 (with 1 being highest toxicity level) to input text based on certain textual attributes. 
While many years have passed since Hosseini et al. \cite{Hosseini-RP3} adversarial attack, our recent evaluation proves that Perspective is still vulnerable to such attacks. 
Nevertheless, Perspective API is used as the baseline model to evaluate the proposed methods. 

\item \textbf{Perspective-based Negated Testset Generation:} The negation of the semantic meaning of a sentence is a highly complex concept requiring formal logic and textual inference \cite{maccartney-manning-2007-natural}. For the purposes of this paper, negation adversarial attacks are implemented with the addition of “not” in front of highly toxic words and phrases. The used dataset is based on Jigsaw’s private and public testsets, available in \cite{kaggle-0}. Initially, the input sentence is segmented into words, some of which selected based on a profanity filter and 
on Perspective's toxicity scoring. More specifically, Perspective selects the words based on their contribution to the sentence's overall toxicity. 
Finally, the term “not” is added in front of each selected term using a PoS tagger \cite{posTagger-2k11} to guide the process. The resulting negated datasets contains 418 and 382 negated sentences from both the private and public Jigsaw datasets and is used to evaluate the proposed methodologies described in section 4.

\item \textbf{LSTM + CNN:} Zhang et. al. \cite{lstm_cnn} propose a combination of LSTM and CNN models to capture both long term dependencies among sentence terms and recognize high-level sentence features. More specifically, an LSTM model receives the initial text sequence and extracts the features describing the dependencies among its words. Then, a CNN model receives the LSTM output feature-set and extracts its high level characteristics. 
In the current model, we use fastText pre-trained word embeddings \cite{fastText} which contain 1 million vectorized data from Wikipedia and other sources. We train the aforementioned hybrid model using Jigsaw’s unintended bias training set \cite{kaggle-0} for consistency during testing with Perspective API. The dataset is split into 80\% for training and 20\% devoted to validation.

\item \textbf{BERT embeddings + Bidirectional LSTM:} BERT \cite{Devlin2019BERTPO} is a model that produces word-level and sentence-level embedding vectors from input text sentences. 
In particular, it is capable of predicting sequences of words and sentences due to the integration of masked-language modeling. 
In the current paper, we use the uncased model as backbone for training a bidirectional LSTM model with extracted BERT embedding representations. A bidirectional LSTM is capable of preserving the reciprocal data-flow from input streams, thus, deducing past and future dependencies. 
The produced embedding dataset was split into 80\% for training and 20\% for validation. The model has 2 bidirectional LSTM layers with 128 LSTM units each, 2 dense hidden layers with 512 units each and was trained with binary cross-entropy loss function. Additionally, a spatial 1D dropout layer is added to drop 1D feature maps instead of individual elements with a 0.3 rate. 
\end{itemize}

\subsection{Results on Hosseini's \cite{Hosseini-RP3} Adversarial Set}
Hosseini's \cite{Hosseini-RP3} adversarial texts are used for initial evaluation of the presented methodologies. More specifically, it consists of 9 highly toxic sentences to which negation was inserted in front of specific profane words. The goal is to calculate an estimate of their capabilities and accuracy. The models are Perspective API, LSTM + CNN and Bert + Bi-LSTM. The results of Table 1 indicate that Perspective API is the model with the most reliable and consistent performance, correctly classifying each of the 9 sentences as toxic. On the contrary, the other 2 models are prone to mistakes missing multiple of the obvious toxic comments, even though both of them were trained using the same training set and reaching high validation scores. Therefore, it is safe to assume that these models should also underperform during the adversarial testing, confusing toxic sentences for non-toxic and vice versa and making their results unreliable. 

\begin{table}[t]
\centering
\begin{tabular}{ |p{2.7cm}||p{1.4cm}|p{1.4cm}|p{1.3cm}|  }
 \hline
 \multicolumn{4}{|c|}{Table 1: Hoessini's \cite{Hosseini-RP3} Original Comments} \\
 \hline
 Original Comment& Perspective Toxicity & LSTM+CNN Toxicity & BERT+Bi-LSTM\\
 \hline
 Climate change is happening and it is not changing in our favor. If you think differently you are an idiot &0.952  &0.66 &0.861 \\
 \hline
 They are stupid, it is getting warmer, we should enjoy it while it lasts &0.873  &0.66 &0.074\\
 \hline
 They are liberal idiots who are uneducated. &0.97  &0.19 &0.893 \\
 \hline
 idiots. backward thinking people. nationalists. not accepting facts. susceptible to lies. &0.949   &0.07 &0.095\\
 \hline
 They are stupid and ignorant with no class &0.949  &0.64 &0.888 \\
 \hline
 It is stupid and wrong &0.952  &0.7 &0.861\\
 \hline
 If they voted for Hilary they are idiots &0.951  &0.01 &0.877 \\
 \hline
 Anyone who voted for Trump is a moron &0.959  & 0.64 &0.368 \\
 \hline
 Screw you trump supporters &0.813 &0.33 &0.068\\
 \hline
\end{tabular} 
\label{tab:1}
\end{table} 

More specifically, the evaluated methodologies are the 1st, 2nd, 3rd, 4th and 5th versions of reasoning and synthesis combined with Perspective API, which are notated using M1.1-P, M1.2-P, M1.3-P, M1.4-P and M1.5-P respectively. 
The 1st and 2nd versions of substitution and exploration method are integrated with Perspective API and named M2.1-P and M2.2-P. The aforementioned formal reasoning wrappers are also combined with LSTM+CNN model and are referenced as M1.1-LC, M1.2-LC and etc. Similarly, they are combined with Bert + Bi-LSTM and are named  M1.1-BL, M1.2-BL and etc. The proposed methods are evaluated against each of the 9 negated toxic sentences presented in Hosseini’s \cite{Hosseini-RP3} second table and tested for 3 different metrics. These metrics are toxicity, severe toxicity and obscene toxicity. We only focus on the plain toxicity metric, for this paper, the results from the other 2 metrics are included for reference and completeness only. In order for a comment to be characterized as non-toxic its toxicity score must be below the 0.5 threshold. 

\begin{table}[t]
\centering
\begin{tabular}{ |p{2.2cm}||p{1.5cm}|p{1.5cm}|p{1.5cm}|  }
 \hline
 \multicolumn{4}{|c|}{Table 2: Hosseini's 1st Negated Sentence} \\
 \hline
 Method& Toxicity &Severe Toxicity&Obscene Toxicity \\
 \hline
 \textbf{Perspective}   & 0.8    &0.335&   0.215\\
 M1.1-P&   0.29  & 0.569   &0.665\\
 M1.2-P &0.107 & 0.086&  0.855\\
 M1.3-P    &0.198 & 0.088&  0.063\\
 M1.4-P    &0.18 & 0.596&  0.027\\
 M1.5-P    &0.17 & 0.596&  0.028\\
 M2.1-P  & 0.265 &0.125 &0.047\\
 M2.2-P& 0.237  & 0.088   &0.022\\
 \textbf{LSTM + CNN}& 0.75  &0.03 &0.147\\
 M1.1-LC& 0.177  & 0.962&0.805\\
 M1.2-LC& 0.027  & 0.114&0.096\\
 M1.3-LC& 0.11  & 0.026&0.004\\
 M1.4-LC& 0.14  & 0.737&0.6\\
 M1.5-LC& 0.141  & 0.7416&0.6\\
 M2.1-LC& 0.14  & 0.006&0.021\\
 M2.2-LC& 0.107  & 0.0&0.0\\
 \textbf{Bert+Bi-LSTM}  & 0.806 &0.025 &0.151\\
 M1.1-BL& 0.21  & 0.946   &0.673\\
 M1.2-BL& 0.07  & 0.101&0.077\\
 M1.3-BL& 0.157  & 0.044&0.014\\
 M1.4-BL& 0.127  & 0.49&0.417\\
 M1.5-BL& 0.1123  & 0.5&0.42\\
 M2.1-BL& 0.144  & 0.004&0.009\\
 M2.2-BL& 0.109  & 0.004&0.021\\
 \hline
\end{tabular} 
\label{tab:2}
\end{table}

The results for the 1st negated sentence are illustrated in table 2, while the the rest of the negated sentences are omitted due to lack of available space. 
 We conclude that methods M1.2 and M1.5 output the best and most reliable performance in comparison to any other heuristic wrapper and backend machine-learning toxicity models. Their success is attributed to the use logic negation alongside redistribution of the scores. 
 M1.2 and M1.5 methods' difference is that the former estimates each phrase's weight based on wordcount, while the latter retrieves an approximation of the weights used by Perspective API by querying it repeatedly. M2.1 and M2.2 generate comparable and average results. However, they are also computationally demanding operations and cannot be used for real-time toxicity detection in online social media platforms. M1.3 is also disqualified despite its good performance, since it constitutes plain toxicity redistribution based on weight determined by word-count. 
In general, Perspective outperforms the other two algorithms (LSTM + CNN and Bert + Bi-LSTM), 
by showcasing reliable and consistent results, while the other models often predict incorrect toxicity scores for the original 9 sentences (table 1). More specifically, they fail to capture the underlying embedded toxicity appropriately, thus, resulting in either too hight or too low toxicity scores for certain cases.

\subsection{Results on Perspective-based Negated Testset}
\label{ss:NegTestSet}

The aforementioned testset contains profane words with the adverb “not” in front of them, however, toxicity doesn't come only from profane words. Therefore, 
the Perspective-based negated public and private datasets is created (discussed in subsection 5.1). More specifically, we evaluate only the most reliable and best performing methodologies from the previous subsection, which are M1.1-P, M1.2-P, M1.4-P and M1.5-P
Additionally, variations of methods M2.1 and M2.2 that discard the paraphrasing modules are tested for completeness. 
Through these modifications, crucial processing steps for adversarial toxicity detection are traded for 
less demanding computational complexity, hypothesizing that their accuracy will still be good enough. All methods utilize Perspective API as their toxicity detector. The evaluation results from the Perspective-based negated adversarial dataset are illustrated in Table 3. 
The accuracy metric corresponds to the number of negated comments recognized as non-toxic with respect to the total number of comments in the testset. The results demonstrate that M1.2 and M1.5 achieve the best performance with respect to the other methodologies, with accuracy scores of 84\% and 82\% respectively. This marks an improvement of over 10\% for the accuracy scores of both methodologies, compared to their performance on the profanity-based negated testset.

\begin{table}[t]
\centering
\begin{tabular}{ |p{2.2cm}||p{1.5cm}|p{2cm}|  }
 \hline
 \multicolumn{3}{|c|}{Table 3:  Perspective-based Negated Public Testset} \\
 \hline
 Method& Accuracy &Marginal Improvement \\
 \hline
 \textbf{Perspective}   & 0.003    &-\\
 M1.1-P&   0.58  & -\\
 M1.2-P &0.82 & 0.92\\
 M1.4-P    &0.72 & -\\
 M1.5-P    &0.84 & 0.90\\
 M2.1-P    &0.42 & 0.48\\
 M2.2-P    &0.37 & 0.47\\
 \hline
\end{tabular} 
\label{tab:13}
\end{table}

\begin{table}[t]
\centering
\begin{tabular}{ |p{2.2cm}||p{1.5cm}|p{2cm}|  }
 \hline
 \multicolumn{3}{|c|}{Table 4:  Perspective-based Negated Private Testset} \\
 \hline
 Method& Accuracy &Marginal Improvement \\
 \hline
 \textbf{Perspective}   & 0.003    &-\\
 M1.1-P&   0.59  & -\\
 M1.2-P &0.87 & 0.92\\
 M1.4-P    &0.76 & -\\
 M1.5-P    &0.86 & 0.87\\
 M2.1-P    &0.40 & 0.52\\
 M2.2-P    &0.40 & 0.49\\
 \hline 
\end{tabular} 
\label{tab:14}
\end{table}

\begin{figure}[ht]
   \centering
    \includegraphics[width=0.45\textwidth]{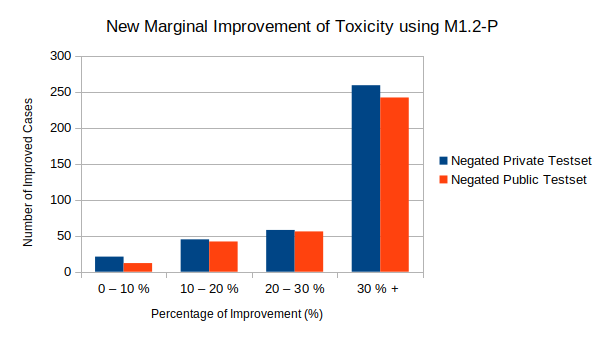}
    \caption{Percentage of improvement of toxicity in the Perspective-based negated public and private test sets using the Method M1.2-P.}
    \label{fig:NewPublicImprovM12}
\end{figure}
\begin{figure}[ht]
   \centering
    \includegraphics[width=0.45\textwidth]{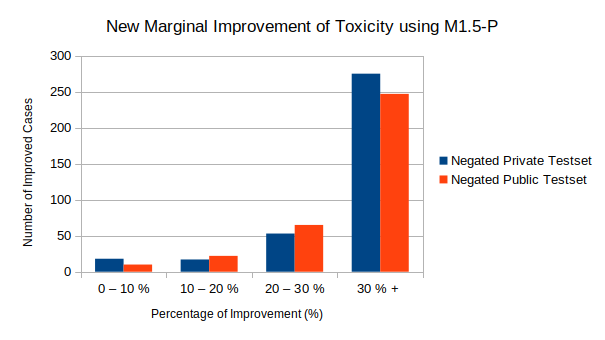}
    \caption{Percentage of improvement of toxicity in the Perspective-based negated public and private test sets using the Method M1.5-P.}
    \label{fig:NewPublicImprovM15}
\end{figure}
\begin{figure}[ht]
   \centering
    \includegraphics[width=0.45\textwidth]{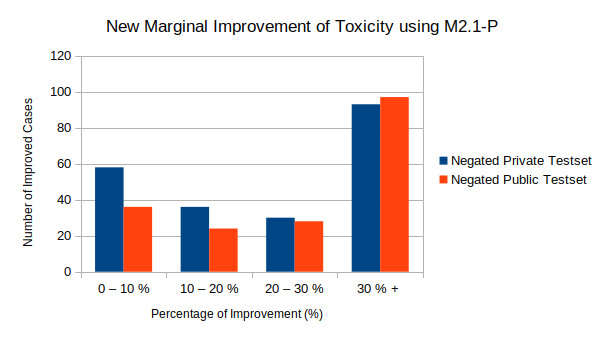}
    \caption{Percentage of improvement of toxicity in the Perspective-based negated public and private test sets using the Method M2.1-P.}
    \label{fig:PublicImprovM21}
\end{figure}
\begin{figure}[ht]
   \centering
    \includegraphics[width=0.45\textwidth]{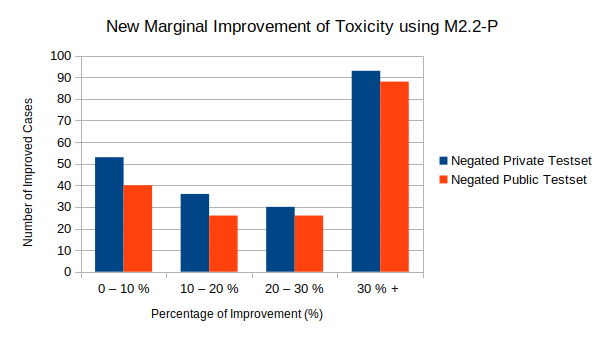}
    \caption{Percentage of improvement of toxicity in the Perspective-based negated public and private test sets using the Method M2.2-P.}
    \label{fig:PublicImprovM22}
\end{figure}

Additionally, both M1.2 and M1.5 showcase a significant gain in reducing the toxicity score of the original sentence. 
Their effectiveness is highlighted even further in figure \ref{fig:NewPublicImprovM12} and figure \ref{fig:NewPublicImprovM15} respectively, which illustrate an over 30\% decrease in the toxicity score of the majority of the input comments. In particular,both the M1.5-P and M1.2-P methodologies provide an improvement in the toxicity of almost 250 out of 382 sentences, in the case of 30\% decrease. Figures \ref{fig:PublicImprovM21} and \ref{fig:PublicImprovM22} 
illustrate the results on the marginal improvement of the M2.1-P and M2.2-P methodologies, which aren't as promising.
Finally, Table 4 contains the results from the evaluation of the same methods against the Perspective-based negated adversarial dataset generated from Jigsaw’s private testset \cite{kaggle-0}. Both of the methodologies M1.2 and M1.5 perform well, with the former achieving an overall accuracy 87\% . As illustrated in figures \ref{fig:NewPublicImprovM12} and figure \ref{fig:NewPublicImprovM15}, both methods also contribute to a significant improvement of the overall toxicity score of each comment. In particular, both M1.2-P and M1.5-P assist in the improvement of the toxicity for more than 250 comments out 418 demonstrate, when achieving a drop of over 30\% from their original toxicity. 

%% file: sections/06-Discussion.tex
\section{Discussion}
\label{s:discuss}

\subsection{Discussion on Results of Evaluation}


The Perspective-based negated adversarial tests highlight the strengths of hybrid methodologies, but methods M2.1 and M2.2 still require improvement. In the comment \textit {“Eat shit you stupid fuk”} from the testset, Perspective API detects the toxic words of the sentence and they are appropriately negated. However, Thesaurus-based methods retrieve antonyms for toxic words, which don't exist for "shit" and "fuk", thus, decreasing the accuracy of the corresponding methods. Performance issues are also attributed to Perspective's behavior and the testset generation method. More specifically, the second adversarial testset relies on detecting words that increase a sentence's toxicity score using the Perspective API. The exact procedure for the computation of a toxicity score within Perspective remains unknown, thus, we hypothesize that even though multiple toxic words may exist in a given sentence, only some of have a big impact on the actual toxicity score. In the comment \textit {“These scumbags need ...Don't drop the soap.”} from the private Jigsaw dataset, the Perspective-based testset generation methodology identifies the words “scumbags” and “rot” as those with the larger impact, but many words remain that are highly toxic. In particular, the combined toxicity of “scumbags” and “rot” is 0.75, which is equal to the sentence's overall toxicity. 

\subsection{Discussion on Performance of Methods}
The results single out the reasoning and synthesis methods M1.2 and M1.5 for their reliability and effectiveness. The substitution and exploration (SE) methodologies are also characterized by consistency, particularly the version combined with the Lesk algorithm \cite{lesk_algo}. However, this is also the most computationally costly solution, which raises the question of balancing between accuracy and efficiency. The complexity of the SE method is the combination of complexities from its individual processing steps. These are (a) retrieval of corresponding antonyms, (b) generation of permutations, (c) paraphrasing for produced permutations and (d) query the toxicity detector. The first process depends on the number of negated toxic words, while the retrieval of the antonyms requires a constant time c with each query. Thus, it can be expressed as \(O(c*n)\).
The second process's complexity is expressed as the number of permuted sentences generated from the retrieved antonyms of negated words. A sentence is generated for each combination of antonyms, therefore the complexity is \(O(5^n)\). The paraphraser produces 10 paraphrases for each sentence, with the scoring process requiring 1 second interval-time between queries. Thus, its complexity is a expressed as \(O(5^n)*10 * 1)\). The SE method's computational complexity is detailed below. \newline

\begin{math}
\label{eqn:heur1}
SE_{Complexity} = O(AntonymRetrieval) + O(Permutations) + O(Paraphrasing) + O(QueringToxicityDetector) \Rightarrow\newline
SE_{Complexity} = O(c*n) + O(5^n) + O((5^n) * 10) + O(5^n *10 * 1) \Rightarrow \newline
SE_{Complexity} = O(n) + O(5^n) + O(5^n) + O(5^n) \Rightarrow \newline
SE_{Complexity} = O(5^n)
\end{math} 



%% file: sections/08-Conclusions.tex
\section{Conclusions \& Future Work}
\label{s:conclude}


Methods protecting against with adversarial text attacks are important to deal with cyberbullying and toxicity in social media platforms or online chatrooms. For this purpose, we propose several hybrid methods combining formal reasoning with existing learning-models for toxicity detection in sentences. 
More specifically, two main approaches and their variations are studied. The first method reverses the toxicity scores of the negated phrases in the comment and normalizes the overall score using coefficients. The second approach reverses the negated phrase by replacing its words with their antonyms. 
The experiments prove the effectiveness of synergistic methods combining features of both formal reasoning and statistical ML to address difficult adversarial attacks. Additionally, Perspective is proven to perform better for toxicity detection than other state-of-the-art methodologies. 
In the future, we will integrate a neologism analysis layer into the existing methodology. More specifically, experiments with various sentences containing neologisms indicate that Perspective predicts high toxicity scores for words with unknown roots. 

%% file: references.bib
@article{Rodriguez2018ShieldingGL,
  title={Shielding Google's language toxicity model against adversarial attacks},
  author={Nestor Rodriguez and Sergio Rojas Galeano},
  journal={ArXiv},
  year={2018},
  volume={abs/1801.01828}
}

@inproceedings{jigsaw-0,
author = {Aroyo, Lora and Dixon, Lucas and Thain, Nithum and Redfield, Olivia and Rosen, Rachel},
title = {Crowdsourcing Subjective Tasks: The Case Study of Understanding Toxicity in Online Discussions},
year = {2019},
isbn = {9781450366755},
publisher = {Association for Computing Machinery},
address = {New York, NY, USA},
url = {https://doi.org/10.1145/3308560.3317083},
doi = {10.1145/3308560.3317083},
booktitle = {Companion Proceedings of The 2019 World Wide Web Conference},
pages = {1100–1105},
numpages = {6},
location = {San Francisco, USA},
series = {WWW '19}
}

@article{BrassardGourdeau2018ImpactOS,
  title={Impact of Sentiment Detection to Recognize Toxic and Subversive Online Comments},
  author={{\'E}loi Brassard-Gourdeau and Richard Khoury},
  journal={ArXiv},
  year={2018},
  volume={abs/1812.01704}
}

@inproceedings{carl-wag,
author = {Carlini, Nicholas and Wagner, David},
year = {2017},
month = {05},
pages = {39-57},
title = {Towards Evaluating the Robustness of Neural Networks},
doi = {10.1109/SP.2017.49}
}

@INPROCEEDINGS{adv_text,
  author={Jain, Edwin and Brown, Stephan and Chen, Jeffery and Neaton, Erin and Baidas, Mohammad and Dong, Ziqian and Gu, Huanying and Artan, Nabi Sertac},
  booktitle={2018 International Conference on Computational Science and Computational Intelligence (CSCI)}, 
  title={Adversarial Text Generation for Google's Perspective API}, 
  year={2018},
  volume={},
  number={},
  pages={1136-1141},
  doi={10.1109/CSCI46756.2018.00220}}

@inproceedings{alllove,
author = {Gr\"{o}ndahl, Tommi and Pajola, Luca and Juuti, Mika and Conti, Mauro and Asokan, N.},
title = {All You Need is "Love": Evading Hate Speech Detection},
year = {2018},
isbn = {9781450360043},
publisher = {Association for Computing Machinery},
address = {New York, NY, USA},
url = {https://doi.org/10.1145/3270101.3270103},
doi = {10.1145/3270101.3270103},
booktitle = {Proceedings of the 11th ACM Workshop on Artificial Intelligence and Security},
pages = {2–12},
numpages = {11},
location = {Toronto, Canada},
series = {AISec '18}
}

@inproceedings{lesk_algo,
author = {Lesk, Michael},
title = {Automatic Sense Disambiguation Using Machine Readable Dictionaries: How to Tell a Pine Cone from an Ice Cream Cone},
year = {1986},
isbn = {0897912241},
publisher = {Association for Computing Machinery},
address = {New York, NY, USA},
url = {https://doi.org/10.1145/318723.318728},
doi = {10.1145/318723.318728},
booktitle = {Proceedings of the 5th Annual International Conference on Systems Documentation},
pages = {24–26},
numpages = {3},
location = {Toronto, Ontario, Canada},
series = {SIGDOC '86}
}

@article{hochreiter1997long,
  title={Long short-term memory},
  author={Hochreiter, Sepp and Schmidhuber, J{\"u}rgen},
  journal={Neural computation},
  volume={9},
  number={8},
  pages={1735--1780},
  year={1997},
  publisher={MIT Press}
}

@inproceedings{Devlin2019BERTPO,
  title={BERT: Pre-training of Deep Bidirectional Transformers for Language Understanding},
  author={Jacob Devlin and Ming-Wei Chang and Kenton Lee and Kristina Toutanova},
  booktitle={NAACL},
  year={2019}
}

@inproceedings{maccartney-manning-2007-natural,
    title = "Natural Logic for Textual Inference",
    author = "MacCartney, Bill  and
      Manning, Christopher D.",
    booktitle = "Proceedings of the {ACL}-{PASCAL} Workshop on Textual Entailment and Paraphrasing",
    month = jun,
    year = "2007",
    address = "Prague",
    publisher = "Association for Computational Linguistics",
    url = "https://aclanthology.org/W07-1431",
    pages = "193--200",
}

@INPROCEEDINGS{lstm_cnn,
  author={Zhang, Jiarui and Li, Yingxiang and Tian, Juan and Li, Tongyan},
  booktitle={2018 IEEE 3rd Advanced Information Technology, Electronic and Automation Control Conference (IAEAC)}, 
  title={LSTM-CNN Hybrid Model for Text Classification}, 
  year={2018},
  volume={},
  number={},
  pages={1675-1680},
  doi={10.1109/IAEAC.2018.8577620}}

@misc{prithivida2021parrot,
  author       = {Prithiviraj Damodaran},
  title        = {Parrot: Paraphrase generation for NLU.},
  year         = 2021,
  version      = {v1.0}
}

@misc{fastText,
  url = {https://fasttext.cc/docs/en/english-vectors.html},
  title = {fastText},
  publisher = {fastText},
  year = {2022},
  note = {Last accessed 27 April 2022}
}

@misc{mayo-0,
  url = {https://www.mayoclinic.org/healthy-lifestyle/tween-and-teen-health/in-depth/teens-and-social-media-use/art-20474437},
  author = {Mayo Clinic Staff},
  title = {Teens and social media use: What's the impact?},
  publisher = {Mayo Clinic},
  year = {2022},
  note = {Last accessed 27 April 2022} 
}

@misc{kaggle-0,
  url = {https://www.kaggle.com/c/jigsaw-unintended-bias-in-toxicity-classification},
  author = {Jigsaw Team},
  title = {Jigsaw Unintended Bias in Toxicity Classification},
  publisher = {Kaggle},
  year = {2022},
  note = {Last accessed 27 April 2022} 
}

@misc{etactics-0,
  url = {https://etactics.com/blog/social-media-and-mental-health-statistics},
  author = {Maria Clark},
  title = {40+ Frightening Social Media and Mental Health Statistics},
  publisher = {etactics},
  year = {2020}
}

@misc{security-0,
  url = {https://www.security.org/resources/cyberbullying-facts-statistics/},
  author = {Security.org Team},
  title = {Cyberbullying: Twenty Crucial Statistics for 2022},
  publisher = {security.org},
  year = {2022}
}

@misc{dataprot-0,
  url = {https://dataprot.net/statistics/cyberbullying-statistics/},
  author = {Ivana Vojinovic},
  title = {Heart-Breaking Cyberbullying Statistics for 2022},
  publisher = {DataProt},
  year = {2022}
}

@misc{entrepreneur-0,
  url = {https://www.entrepreneur.com/article/328749},
  author = {Tobin Brogunier},
  title = {4 Reasons Why Social Media Has Become So Toxic and What to Look for Next},
  publisher = {Entrepreneur},
  year = {2019}
}

@misc{WIRED-Greenberg,
  url = {https://www.wired.com/2017/02/googles-troll-fighting-ai-now-belongs-world/},
  author = {Andy Greenberg},
  title = {Now Anyone Can Deploy Google's Troll-Fighting AI},
  publisher = {WIRED},
  year = {2017}
}

@misc{TheRegister-Claburn,
  url = {https://www.theregister.com/2017/03/02/google\_trollspotting\_ai\_trips\_over\_typos/},
  author = {Thomas Claburn},
  title = {Google's troll-destroying AI can't cope with typos},
  publisher = {TheRegister},
  year = {2017}
}

@misc{Hosseini-RP3,
  doi = {10.48550/ARXIV.1702.08138},
  url = {https://arxiv.org/abs/1702.08138},
  author = {Hosseini, Hossein and Kannan, Sreeram and Zhang, Baosen and Poovendran, Radha},
  title = {Deceiving Google's Perspective API Built for Detecting Toxic Comments},
  publisher = {arXiv},
  year = {2017},
  copyright = {arXiv.org perpetual, non-exclusive license}
}

@article{L2R1,
  title={Human Decision Making Model for Autonomic Cyber Systems},
  author={Mertoguno, J. Sukarno},
  journal={International Journal of Artificial Intelligence Tools},
  volume={23},
  number={6},
  pages={1--6},
  year={2014},
  publisher={World Scientific}
}

@InProceedings{posTagger-2k11,
author="Manning, Christopher D.",
editor="Gelbukh, Alexander F.",
title="Part-of-Speech Tagging from 97{\%} to 100{\%}: Is It Time for Some Linguistics?",
booktitle="Computational Linguistics and Intelligent Text Processing",
year="2011",
publisher="Springer Berlin Heidelberg",
address="Berlin, Heidelberg",
pages="171--189",
isbn="978-3-642-19400-9"
}

@inproceedings{NEURIPS2020_adaptive_attacks,
 author = {Tramer, Florian and Carlini, Nicholas and Brendel, Wieland and Madry, Aleksander},
 booktitle = {Advances in Neural Information Processing Systems},
 editor = {H. Larochelle and M. Ranzato and R. Hadsell and M.F. Balcan and H. Lin},
 pages = {1633--1645},
 publisher = {Curran Associates, Inc.},
 title = { On Adaptive Attacks to Adversarial Example Defenses},
 url = {https://proceedings.neurips.cc/paper/2020/file/11f38f8ecd71867b42433548d1078e38-Paper.pdf},
 volume = {33},
 year = {2020}
}

@article{ICLR_adversar,
author = {Madry, Aleksander and Makelov, Aleksandar and Schmidt, Ludwig and Tsipras, Dimitris and Vladu, Adrian},
year = {2017},
month = {06},
pages = {},
title = {Towards Deep Learning Models Resistant to Adversarial Attacks}
}
